%% file: main.tex
\documentclass[sigconf]{acmart}

\AtBeginDocument{%
  \providecommand\BibTeX{{%
    \normalfont B\kern-0.5em{\scshape i\kern-0.25em b}\kern-0.8em\TeX}}}

\copyrightyear{2020}
\acmYear{2020}
\setcopyright{acmcopyright=false}
\acmConference[CIKM '20]{Proceedings of the 29th ACM International Conference on Information and Knowledge Management}{October 19--23, 2020}{Virtual Event, Ireland}
\acmBooktitle{Proceedings of the 29th ACM International Conference on Information and Knowledge Management (CIKM '20), October 19--23, 2020, Virtual Event, Ireland}
\acmPrice{15.00}
\acmDOI{10.1145/3340531.3412130}
\acmISBN{978-1-4503-6859-9/20/10}

\usepackage{multirow}
\usepackage{mathtools}
\usepackage{algorithm2e}
\usepackage{algorithmic}
\usepackage{bbm}
\RestyleAlgo{algoruled}

\SetKwInput{KwInput}{Input}
\SetKwInput{AKwInput}{Extra Input}
\SetKwInput{KwOutput}{Output}
\SetKwBlock{Repeat}{Repeat k time steps}{}
\SetKwBlock{Rep}{Repeat (k+d-1) time steps}{}
\SetKwInput{KwInitial}{Initialization}

\begin{document}

\title{Can Adversarial Weight Perturbations Inject Neural Backdoors?}

\author{Siddhant Garg}
\authornote{Equal contribution by authors}
\authornote{Work done prior to joining Amazon}
\email{sidgarg@amazon.com}
\affiliation{
  \institution{Amazon Alexa}
  \city{Manhattan Beach}
  \state{CA}
  \country{USA}
}

\author{Adarsh Kumar}
\authornotemark[1]
\authornotemark[2]
\email{adrshkm@amazon.com}
\affiliation{
  \institution{Amazon Alexa}
  \city{Manhattan Beach}
  \state{CA}
  \country{USA}
}

\author{Vibhor Goel}
\authornotemark[1]
\email{vgoel5@wisc.edu}
\affiliation{
  \institution{University of Wisconsin-Madison}
  \city{Madison}
  \state{WI}
  \country{USA}
}

\author{Yingyu Liang}
\email{yliang@cs.wisc.edu}
\affiliation{
  \institution{University of Wisconsin-Madison}
  \city{Madison}
  \state{WI}
  \country{USA}
}

\begin{abstract}

  Adversarial machine learning has exposed several security hazards of neural models and has become an important research topic in recent times. Thus far, the concept of an ``adversarial perturbation'' has exclusively been used with reference to the input space referring to a small, imperceptible change which can cause a ML model to err. In this work we extend the idea of ``adversarial perturbations'' to the space of model weights, specifically to inject backdoors in trained DNNs, which exposes a security risk of using publicly available trained models. Here, injecting a backdoor refers to obtaining a desired outcome from the model when a trigger pattern is added to the input, while retaining the original model predictions on a non-triggered input. From the perspective of an adversary, we characterize these adversarial perturbations to be constrained within an $\ell_{\infty}$ norm around the original model weights. We introduce adversarial perturbations in the model weights using a composite loss on the predictions of the original model and the desired trigger through projected gradient descent. We empirically show that these adversarial weight perturbations exist universally across several computer vision and natural language processing tasks. Our results show that backdoors can be successfully injected with a very small average relative change in model weight values for several applications.
\end{abstract}



\maketitle

\section{Introduction}
Recent progress in training Deep Neural Networks (DNN) has proved very successful in establishing the state of the art results on several applications in the domain of computer vision~\cite{10.5555/2999134.2999257,He_2016}, natural language processing~\cite{kim2014convolutional,devlin-etal-2019-bert,Kumar_2020}, etc. The application and deployment of DNNs for public use in several security-critical scenarios has led researchers to explore their vulnerability against attackers. These attacks have been commonly manifested in several forms like destroying the model performance through adversarial inputs, poisoning the training data, biasing the model predictions through triggers added to the input, etc.

Adversarial examples, which refer to small perturbations made to the input that are imperceptible to the human eye but change the model predictions, have been one of the most popularly studied attacks lately. Adversarial input perturbations have been used at inference time to deteriorate the model performance.

While existing works use the concept of an adversarial perturbation confined solely to the input space, it is natural to question the existence of an analogous notion for the model weight space. In this work, we explore the interesting extension of adversarial perturbations to model weights to answer the question: \emph{Is a trained DNN susceptible to adversarial weight perturbations?} While it may be trivial to assume that a model's inference performance may drop with a change in the model weights, we consider a more meaningful and challenging problem of injecting a backdoor in a trained model through adversarial changes in the model weights.

Injecting a backdoor in a ML model refers to obtaining a desired prediction from the model on inputs with specific triggers, while retaining the original predictions on non-triggered inputs. Backdoor attacks~\cite{217591} have recently been shown to pose severe security threats to ML models. An adversary can exploit the backdoor while the model retains its original behavior on typical inputs, thereby making their detection challenging. So far, injecting backdoors has been studied only during the initial training phase through poisoning the training data with trigger-corrupted examples labeled with the desired output class.

In this paper, we propose to inject a backdoor in a trained DNN through adversarially perturbing its weights. Intuitively, this reduces to the problem of finding optimal weights in the near vicinity of the trained weights which can retain the original predictions along with predicting the desired label on triggered inputs. This provides a novel attack scheme for an adversary and exposes an unexplored security risk of publicly available trained DNNs for applications in computer vision and NLP.

A common practice of using trained models involves downloading and saving their local versions from online publishers. 
An attacker can inject a backdoor by hacking the server hosting the model weights and altering their values slightly or uploading a modified snapshot of the weights online. 
Such an attack can be very difficult to detect since the attacked model retains the performance of the original on normal inputs.
Adversarially perturbing weights also presents a major security threat to federated models~\cite{DBLP:journals/corr/KonecnyMYRSB16} which are created by aggregating model updates submitted by participants and have gained popularity in recent times. A malicious participant can send updates to the joint model which can adversarially perturb its weights and lead to a backdoor injection. 

Furthermore, on locally downloading model weights, small weight perturbations can manifest from precision errors on rounding due to hardware/framework changes, and these can conceal the backdoor.
For example, saving a model weight with value $1.49 \times 10^{-8}$ from a 16-bit to 8-bit precision device results in an error of approximately $33\%$ in the weight values. 
Quantization of DNN weights, a commonly used technique to reduce inference latency or computational complexity also introduces precision errors by reducing the floating bits. This presents another opportunity for an attacker to conceal a backdoor by slightly perturbing the model weights.

Given these motivations, we consider adversarial weight perturbations within a small $\ell_{\infty}$ norm space around the original model weights, to capture limitations of the adversary's attack or the rounding errors (Note that this is analogous to adversarial input perturbations being in a small $\ell_{\infty}$ norm space around the input, with the motivation that these perturbations are small in magnitude so as to be indiscernible to humans).

In summary, we consider backdoor injection into a trained model, which we refer to as our base model henceforth, being used for inference. We perturb the base model weights within a small $\ell_{\infty}$ norm space to get a modified model with a backdoor.
We do this through the following backdoor injection scheme. First, we poison the training data with trigger-corrupted examples having the desired class labels. Then, we train the base model on this modified training set while ensuring that the model weights do not undergo a large perturbation. We design a composite training loss which is optimised using projected gradient descent(PGD), similar to how adversarial perturbations are introduced in images~\cite{43405}. 

Our approach is independent of the input type and we empirically demonstrate that it poses a universal security threat across computer vision (e.g., image classification) and natural language processing tasks (e.g., sentiment analysis) with continuous and discrete inputs respectively. Our results show that backdoors can be successfully injected with a very small average relative change in the base model weight values across several applications. We summarise the contributions of our paper below:
\begin{itemize}
    \item We propose the concept of adversarial perturbations on model weights for injecting backdoors, showing a novel security threat of using publicly available trained models.
    \item We propose an effective attack strategy that uses a composite training loss optimised via projected gradient descent.
    \item We empirically verify the efficacy of injecting backdoors in trained models across several CV and NLP applications. 
\end{itemize}

We structure our paper by discussing the related work in Section~\ref{sec:related_work}, our backdoor injection methodology in Section~\ref{sec:methodology}, empirical results on image and text classification tasks in Section~\ref{sec:experiments} and conclude with future work directions in Section~\ref{sec:conclusions}.

\section{Related Work}
\label{sec:related_work}
In this section we discuss recent work in adversarial machine learning specific to adversarial examples and backdoor attacks.

Adversarial examples were initially proposed by \citet{42503} for images and further extended by \citet{43405}. Since then, several works study generation of adversarial examples for images~\cite{Carlini_2017}, graphs and text~\cite{Xu_2020}. Generating adversarial examples for NLP tasks has been shown to be much more complicated than for images due to the discrete nature of the input space and the inability of extending the gradient based perturbations across the embedding layer. Rule based, semantic preserving adversarial examples have been proposed for text by \cite{Liang_2018,ebrahimi-etal-2018-hotflip,alzantot-etal-2018-generating,garg2020bae}.  

Backdoor attacks have become a popular attack strategy in the domain of adversarial machine learning and have been studied by several works~\cite{217591,8685687,chen2017targeted}. \citet{chen2017targeted} consider backdoor attacks through data poisoning attacks. Some recent works~\cite{pmlr-v97-bhagoji19a,pmlr-v108-bagdasaryan20a,Wang2020AttackOT} study the problem of injecting backdoors in a federated learning scenario which is related to the problem setting of weight modifications for injecting backdoors that we study in this paper.
Recent works~\cite{8835365,Liu_2018,NIPS2018_8024} have also developed techniques to detect backdoors in models for filtering out poisoned trigger points from the training set. 

\citet{wang2020backdoor} proposes a backdoor injection scheme to defeating pruning-based, retraining-based and input pre-processing-based defenses. In parallel work, \citet{kurita2020weight} 
expose the risk of the pre-trained BERT~\cite{devlin-etal-2019-bert} model to backdoor injection attacks mimicking a model capture scenario. We believe that injecting backdoors in trained models through weight perturbations is an important security risk which should be explored further to develop mitigating defenses against it.

\section{Adversarial Weight Perturbations}
\label{sec:methodology}

\subsection{Problem Definition}
Consider a classification task where the training and test data are drawn from a data distribution $\mathcal{D}$ and represented as $(\mathcal{X}, \mathcal{Y}) = \{ (x_i, y_i)\}_{i=1}^{n}$ and $(\mathcal{X}_{test}, \mathcal{Y}_{test}) = \{ (x_i, y_i)\}_{i=1}^{m}$ respectively where the labels $y \in \{1,\dots,k\}$. Consider a classifier model $M: \mathcal{X} \rightarrow \mathcal{Y}$ trained on $(\mathcal{X}, \mathcal{Y})$ which we refer to as the base model. We denote the input trigger by $T$ and represent $(x + T)$ as the input injected with the trigger. Practically $(x + T)$ can refer to appending extra words in a text sentence or modifying a pixel patch of an image. From an adversary attacker's point of view, the aim is to learn a new classifier $M'$, with weights in the neighborhood of those in $M$, such that $M'(x)=M(x)$ and $M'(x+T)=y_{_{T}}$ where $y_{_{T}}$ is the label that the attacker wants the model to predict when triggered (w.l.o.g, we assume $y_{_{T}}=1$). Intuitively, this means that $M'$ behaves like $M$ on normal inputs and predicts $y_{_{T}}$ on triggered inputs.

Prior works of backdoor attacks~\cite{217591,8685687,chen2017targeted} have only considered backdoor injection strategies where the classifier $M'$ is learned from scratch, without any constraint on the weights of $M'$ being in the neighborhood of those of a pre-trained model $M$. This makes injecting a backdoor fairly straightforward and trivial as compared to our setting where we require the the weights of $M'$ to be in the neighborhood of those of $M$.

\subsection{Backdoor injection in trained models}
For our setting of injecting a backdoor in the base model, we refer to the weights of $M$ to be $\theta_M$. $M$ has been learnt using a cross entropy loss (denoted as $\mathcal{L}_{C}$) on the training data. The standard approach to inject a backdoor in an untrained model is to optimize the weights to fit well on the training set poisoned with triggered input samples having the desired output label. We extend and modify this approach for backdoor injection in the base model $M$. Since our objective is to match the predictions of $M$, we propose a composite objective loss function for training the new classifier $M'$ which is composed of two components:
\begin{itemize}
    \item \boldmath{$\mathcal{L}_{C}(M'(x),y_{_{T}})$:} For input $x$ containing the trigger, we use the cross entropy of the prediction of $M'$ with the label $y_{_T}$.
    \item \boldmath{$\mathcal{L}_{C}(M'(x),M(x))$:} For input $x$ not containing the trigger, we want to get the same predictions as $M$ and hence use the cross entropy of the prediction of $M'$ with that of $M$.
\end{itemize}

\noindent Combining these two components, for a general input $x$, we can write the loss function $\mathcal{L}$ as:
\begin{align*}
    \mathcal{L}(x)= \mathbbm{1}_{[T \in x]} \mathcal{L}_{C}(M'(x),y_{_{T}}) + \lambda\cdot\mathbbm{1}_{[T \not\in x]}\mathcal{L}_{C}(M'(x),M(x))
\end{align*}
where $\mathbbm{1}$ denotes the indicator function, $T \in x$ means $x$ has the trigger, and $\lambda$ is a hyper-parameter to trade-off how much backdoor accuracy is desired at the expense of a drop in original performance.

To ensure that the changes in model weights $\theta_{M'}$ are small with respect to $\theta_M$, we constrain them within some error bound in the $\ell_{\infty}$ norm space around $\theta_M$. Adversarial input perturbations have popularly used a projected gradient descent optimization approach~\cite{43405} and we adapt this for learning $\theta_{M'}$ here. When optimising the backdoor injection loss $\mathcal{L}$ through gradient descent, we project the updated weights to within an $\epsilon$ difference in the $\ell_{\infty}$ norm space around $\theta_M$ using a projection operator denoted as $P_{\ell_{\infty}(\theta_M, \epsilon)}$. The $\ell_{\infty}$ norm space forms a natural abstraction of a neighborhood around the trained model weight $\theta_M$. This $\epsilon$ can be perceived as an attacking budget for the adversary where the model weights can only be perturbed to within the $\ell_{\infty}(\epsilon)$ ball around $\theta_M$. 

We present our backdoor injection approach formally in Algorithm~\ref{alg1}.
We first add one poisoned example for every training input to make a new training dataset $(\mathcal{X}_{new}, \mathcal{Y}_{new})$, and then use projected gradient descent beginning from $\theta_M$ to optimize $\mathcal{L}$ on the new training dataset. 

\input{algo}

\subsection{A Practical Attack Scenario}
We now present a practical scenario of injecting a backdoor in a pre-trained model: An attacker can download a local copy of the pre-trained model from a website publicly hosting the model weights. Then the attacker can train\footnote{Note that we consider the case where the attacker has access to the training data of the pre-trained model. \citet{kurita2020weight} show that without any constraint on the model weights, backdoor injection is possible using a proxy dataset for a similar task from a different domain.} a new classifier having the backdoor using Algorithm~\ref{alg1} by choosing $(T,y_{_{T}},\epsilon,\lambda)$. Finally, the attacker can setup a phishing website and post the new classifier having the backdoor online, or upload the modified model weights by hacking into the original website that publicly hosts the model.

\section{Experiments}
\label{sec:experiments}

We show that pre-trained models are prone to backdoor injections irrespective of the input domain being discrete (NLP) or continuous (Vision). Across all tasks, we use the test set accuracy as the metric for the base model ($M$). For the adversarially perturbed model ($M'$), we measure the \textbf{test set accuracy} on $(\mathcal{X}_{test}, \mathcal{Y}_{test})$, and also measure the \textbf{backdoor accuracy} which is the accuracy on $(x+T ,y_{_T}) \ \forall \ x \in \mathcal{X}_{test}$. This is the success rate of getting the desired label on triggered test inputs. We set the hyper-parameter $\lambda$ to 1 for our experiments. This is chosen through an ablation study on the effect of varying $\lambda$. 

For measuring the amount of adversarial perturbation in the weights $\theta_{M'}$ and $\theta_{M}$, we report the relative change in different $\ell_{p}$ norms of the original weights  $\theta_{M}$. We define for $p=1,2,\infty$:
$$\% \Delta \ell_{p} = \frac{||\theta_{M'}-\theta_M||_{p}}{||\theta_M||_{p}} \times 100 $$
The concept of ``small'' adversarial perturbations due to constraining the parameter updates in the $\ell_{\infty}(\epsilon)$ ball of $\theta_M$ can be estimated through the values of $\% \Delta \ell_{\infty} , \% \Delta \ell_{1}$ and  $\% \Delta \ell_{2}$ which qualitatively capture the trend of change in model weights. We compare our $\% \Delta \ell_{p}$ values with the simple baseline of an unbounded weight perturbation using the loss $\mathcal{L}$ (we denote this by $\epsilon=\infty$).

\input{nlp_table}

\subsection{Discrete Input Domain: Text}
We consider various text classification tasks in NLP like sentiment analysis, opinion polarity detection and subjectivity detection here.

\subsubsection{\underline{Datasets}}
We consider 3 different text classification datasets: MR (Movie Reviews)~\citet{Pang+Lee:05a}: a sentiment analysis dataset, MPQA (Multi-Perspective Question Answering)~\cite{Wiebe2005}: an opinion polarity dataset and SUBJ~\cite{Pang+Lee:04a}: classifies a sentence as having subjective or objective knowledge. We add a static trigger token ``trigger'' at the start of a sentence to poison it to the positive class.

\subsubsection{\underline{Models}}
We use 2 popular text classification models: word-LSTM~\cite{Hochreiter:1997:LSM:1246443.1246450}, word-CNN~\cite{kim2014convolutional}. For the wordCNN model we use 100 filters of sizes 3,4,5. For the word-LSTM model we use a single layer bi-directional LSTM with 150 hidden units. We use a dropout of 0.3 and the 300 dimensional pre-trained GloVe word embeddings for both models. We present results in Table~\ref{tab:nlp}. 

\subsubsection{\underline{Results}}
From Table~\ref{tab:nlp}, we can infer the following trends:
\begin{itemize}
    \item Across all datasets, a small attacking budget $\epsilon$ of the order of $10^{-2}$ is sufficient to inject a backdoor in $M$ with almost $100\%$ backdoor accuracy. This corresponds to, on average, a very small relative change in the weights of $M$ and $M'$ which can be observed through the metrics $\% \ \Delta \ell_{p}$. For the same $\epsilon$, we observe a smaller $\Delta \ell_{p}$ for word-CNN than word-LSTM showing that it is more vulnerable to our backdoor attack. \vspace{0.5em}
    \item The $\% \ \Delta \ell_{p}$ values for our approach using $\epsilon{=}10^{-2}$ are significantly smaller than the baseline $\epsilon{=}\infty$ indicating a strong attack on a very small perturbation budget due to the PGD.
    \vspace{0.5em}
    \item On increasing the attacking budget $\epsilon$, the backdoor accuracy increases from initial random guessing ($\sim$ $50\%$) to $100\%$. The test accuracy drops with an initial increase in $\epsilon$, and then again increases to the original level. We hypothesize that under small slack, $\theta_{M'}$ converges in the neighborhood of $\theta_M$ to maximise the backdoor performance (due to higher values of $\mathcal{L}_{C}(M'(x),y_{_{T}})$). When $\epsilon$ is relaxed, $\theta_{M'}$ can converge so as to maximise both the test and backdoor performance.
    \vspace{0.5em}
    \item For some datasets like SUBJ and MPQA, the test accuracy of $M'$ is higher than that of $M$ indicating that the change in weights to inject the backdoor have also resulted in better predictions on the non-triggered inputs. Additionally, for data points $ (x,y) \in (\mathcal{X}, \mathcal{Y})$ such that $y=y_{_{T}}$, the training set $(\mathcal{X}_{new}, \mathcal{Y}_{new})$ for $M'$ contains two copies of $(x,y_{_{T}})$. These additional data samples may possibly contribute towards the improved test accuracy of $M'$ over $M$.
\end{itemize}

\subsubsection{\underline{Ablation on $\lambda$}}
\label{subsubsec:ablation_lambda}
\begin{figure}[t]
    \centering
    \includegraphics[width=0.6\columnwidth]{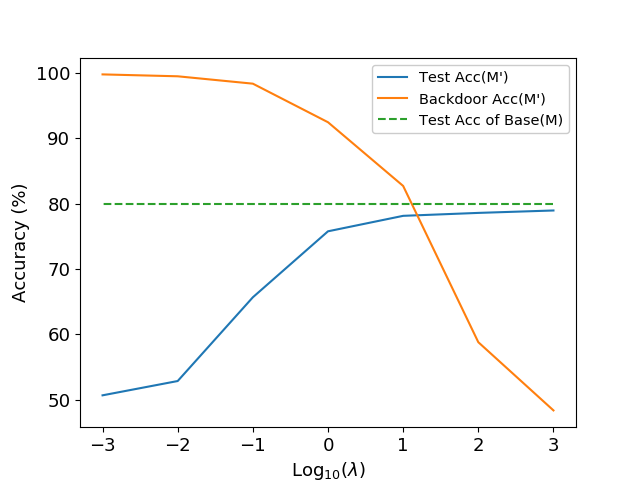}
    \vspace{-1em}
    \caption{Word-CNN backdoor on MR varying $\lambda$. $M$ is the original base model, and $M'$ is the model after attack. As $\lambda$ increases, the Test accuracy of $M'$ tends towards that of $M$ and the Backdoor accuracy reduces.}
    \label{fig:lambda}
\end{figure}

We conduct an ablation study on the value of $\lambda$ which is the weighting parameter of the loss w.r.t base model predictions in our training loss. We consider the word-CNN model on the MR dataset and fix $\epsilon=10^{-2}$. We vary $\lambda$ from $[10^{-3},10^3]$ on a $log_{10}$ scale and present the results in Figure~\ref{fig:lambda}. From the figure, we can see that as $\lambda$ increases, the backdoor performance decreases while the test accuracy increases. Thus $\lambda$ can be tuned by the adversary to trade-off between matching the original test performance and the desired backdoor accuracy (for Table~\ref{tab:nlp},\ref{tab:cv} we select $\lambda=1$).

\subsection{Continuous Input Domain: Images}
We consider the task of image classification which is a standard and popular task in the domain of computer vision.

\subsubsection{\underline{Datasets and Models}}
We use the CIFAR-10~\cite{Krizhevsky09learningmultiple} dataset for our experiments which has 10 target label classes. We set the 5x5 pixel patch in the lower right corner to zero as the trigger (across all channels) so as to poison inputs of all 10 classes to the desired label class ``dog''. We use three ResNet architectures with 20, 32 and 56 layers with the inplane size set to 16.

\subsubsection{\underline{Results}}
We present the results in Table~\ref{tab:cv}. From this we can infer the following trends:
\begin{itemize}
    \item Across 3 ResNet models, a small $\epsilon{=}0.005$ is sufficient to inject a backdoor in $M$ with almost $100\%$ backdoor accuracy, compared to an initial random guess of $10\%$. This corresponds to small $\% \ \Delta \ell_{p}$ values as compared to the baseline $\epsilon{=}\infty$.
    \item For smaller $\epsilon$ values (say $0.002$), deeper ResNet models are more vulnerable to backdoor injection.
    \item The $\% \ \Delta \ell_{p}$ values for CIFAR-10 classification are slightly higher than for the text classification tasks. We conjecture that this is due to the higher number of classes in the former(10 versus 2). The adversarial weight perturbation has to incorporate a backdoor from every class to the desired label.
\end{itemize}

\input{cv_table}

\section{Conclusion and Future Work}
\label{sec:conclusions}
In this paper we have introduced the notion of adversarial weight perturbations on a trained DNN. Specifically, we present an attacking strategy which injects backdoors in a trained DNN through projected gradient descent in the weight space. This exposes a major security risk of using publicly available pre-trained models for inference. Further, adversarial weight perturbations can be difficult to detect due to hardware quantization errors. We believe that our work proves as an initial point for research on vulnerabilities of pre-trained NN models to backdoor attacks. Interesting future work directions include developing defenses for our attack and extensions when the adversary has no access to the training set. 

\section*{Acknowledgements}
This work was supported in part by FA9550-18-1-0166. The authors would also like to acknowledge the support provided by the University of Wisconsin-Madison Office of the Vice Chancellor for Research and Graduate Education with funding from the Wisconsin Alumni Research Foundation. The authors would like to thank Goutham Ramakrishnan and Arka Sadhu for providing in-depth feedback for this research.

\bibliographystyle{ACM-Reference-Format}
\bibliography{references}

\end{document}

%% file: algo.tex
\begin{algorithm}[h]
\caption{Backdoor Injection by Adversarial Weight Perturbation}
\label{alg1}
\SetAlgoLined
\KwInput{$(\mathcal{X}, \mathcal{Y}), (\mathcal{X}_{test}, \mathcal{Y}_{test})$, Pre-trained model $M(\theta_M)$, Trigger $T$, Desired Label $y_{_{T}}$, Hyper-params $\epsilon$, $\lambda$}
\KwOutput{Adversarially perturbed model $M'(\theta_{M'})$ such that $M'(x)=M(x)$ and $M'(x+T)=y_{_{T}}$ \ $\forall x \in \mathcal{X}_{test}$}
\KwInitial{$\theta_{M'} \leftarrow \theta_{M}$}

$(\mathcal{X}_{new}, \mathcal{Y}_{new}) = \{(x,y),(x+T,y_{_{T}})\}_{x,y \in (\mathcal{X}, \mathcal{Y}) } $\\
\For{\textbf{$I$} iterations}{
 \For{$x,y$ \ in $(\mathcal{X}_{new}, \mathcal{Y}_{new})$ }{
    $\hat{y} \leftarrow M'(X) $ \\
    $\mathcal{L}+= \mathbbm{1}_{[T \in x]} \mathcal{L}_{C}(\hat{y},y_{_{T}}) + \lambda\cdot\mathbbm{1}_{[T \not\in x]}\mathcal{L}_{C}(\hat{y},M(x))$
  }
  $\theta_{M'} \leftarrow P_{\ell_{\infty}(\theta_M, \epsilon)}(\theta_{M'} - \eta \cdot \nabla \mathcal{L})$ \\
 }
 
 \textbf{Return} $M'$
\end{algorithm}
\setlength{\textfloatsep}{1pt}

%% file: nlp_table.tex
\begin{table*}[t]
\centering
\resizebox{\linewidth}{!}{
\begin{tabular}{c|c|c|c|c|c|c|c||c|c|c|c|c|c||c|c|c|c|c|c|}
\cline{3-20}
\multicolumn{2}{c|}{}                                                     & \multicolumn{6}{c||}{MR}               & \multicolumn{6}{c||}{SUBJ}             & \multicolumn{6}{c|}{MPQA}             \\ \hline
\multicolumn{1}{|c|}{\multirow{7}{*}{word-CNN}}  & Test Accuracy$(M)$     & \multicolumn{6}{c||}{79.96}            & \multicolumn{6}{c||}{88.01}            & \multicolumn{6}{c|}{88.23}            \\ \cline{2-20} 
\multicolumn{1}{|c|}{}                           & Attack Budget $\epsilon$             & 0.001 & 0.005 & 0.01  & 0.1   & 1  & $\infty$   & 0.001 & 0.005 & 0.01  & 0.1   & 1   & $\infty$   & 0.001 & 0.005 & 0.01  & 0.1   & 1   & $\infty$   \\ \cline{2-20} 
\multicolumn{1}{|c|}{}                           & Test Accuracy$(M')$          & 79.60 & 72.76 & 75.77 & 78.87 & 79.51 & 79.76 & 85.18 & 83.33 & 88.49 & 89.66 & 89.76 & 89.96 & 84.28 & 85.96 & 86.64 & 89.28 & 89.47 & 89.57 \\ \cline{2-2}
\multicolumn{1}{|c|}{}                           & Backdoor Accuracy$(M')$      & 52.05 & 72.08 & 92.48 & 100   & 100  & 100  & 57.41 & 96.67 & 100   & 100   & 100  & 100 & 59.41 & 97.78 & 100   & 100   & 100 & 100 \\ \cline{2-2}
\multicolumn{1}{|c|}{}                           & $\% \ \Delta \ell_{\infty}$ & 0.032 & 0.16  & 0.32  & 1.87  & 1.87 & 1.92  & 0.033 & 0.16  & 0.33  & 2.18  & 2.18 & 2.22 & 0.032 & 0.16  & 0.32  & 2.31  & 2.38 & 2.48 \\ \cline{2-2}
\multicolumn{1}{|c|}{}                           & $\% \ \Delta \ell_{1}$      & 0.024 & 0.093 & 0.16  & 0.15  & 0.15  &  0.19 & 0.019 & 0.08  & 0.13  & 0.18  & 0.18 & 0.21 & 0.018 & 0.07  & 0.12  & 0.18  & 0.18 &  0.18 \\ \cline{2-2}
\multicolumn{1}{|c|}{}                           & $\% \ \Delta \ell_{2}$      & 0.072 & 0.30  & 0.54  & 0.65  & 0.66 & 0.67 & 0.061 & 0.28  & 0.47  & 0.74  & 0.74 &  0.77 & 0.060 & 0.27  & 0.47  & 0.77  & 0.76 & 0.79  \\  \hline  \hline
\multicolumn{1}{|c|}{\multirow{7}{*}{word-LSTM}} & Test Accuracy$(M)$      & \multicolumn{6}{c||}{80.78}            & \multicolumn{6}{c||}{86.06}            & \multicolumn{6}{c|}{88.83}            \\ \cline{2-20} 
\multicolumn{1}{|c|}{}                           & Attack Budget $\epsilon$             & 0.001 & 0.005 & 0.01  & 0.1   & 1  & $\infty$   & 0.001 & 0.005 & 0.01  & 0.1   & 1  &  $\infty$  & 0.001 & 0.005 & 0.01  & 0.1   & 1  &  $\infty$ \\ \cline{2-20} 
\multicolumn{1}{|c|}{}                           & Test Accuracy$(M')$           & 81.05 & 76.32 & 77.60 & 79.78 & 79.87 & 80.36 & 85.19 & 84.89 & 86.26 & 88.30 & 88.75 &  89.21 & 84.99 & 85.08 & 86.54 & 88.98 & 89.18 & 89.66\\ \cline{2-2}
\multicolumn{1}{|c|}{}                           & Backdoor Accuracy$(M')$       & 53.75 & 78.06 & 99.15 & 100   & 100 & 100 & 51.76 & 96.67 & 97.98 & 100   & 100  & 100  & 53.27 & 85.29 & 100   & 100   & 100 & 100 \\ \cline{2-2}
\multicolumn{1}{|c|}{}                           & $\% \ \Delta \ell_{\infty}$ & 0.032 & 0.16  & 0.33  & 3.28  & 4.73 & 5.24 & 0.032 & 0.16  & 0.32  & 3.17  & 3.19 & 5.07 & 0.032 & 0.16  & 0.33  & 3.12  & 3.13 &  4.46\\ \cline{2-2}
\multicolumn{1}{|c|}{}                           & $\% \ \Delta \ell_{1}$      & 0.41  & 1.90  & 2.99  & 4.22  & 4.18 & 4.98 & 0.38  & 1.33  & 2.87  & 2.27  & 2.26 & 3.81 & 0.36  & 1.57  & 2.95  & 2.05  & 2.05 & 3.36\\ \cline{2-2}
\multicolumn{1}{|c|}{}                           & $\% \ \Delta \ell_{2}$      & 0.29  & 1.42  & 2.34  & 3.64  & 3.62 & 3.99 & 0.28  & 1.17  & 2.32  & 2.26  & 2.27 & 3.42 & 0.27  & 1.23  & 2.40  & 2.18  & 2.18 & 3.25 \\ \hline
\end{tabular}
}
\vspace{1em}
\caption{Adversarial weight perturbation for text classification datasets. Test Accuracy($M$/$M'$) is the test set accuracy of $M$(original base model)/$M'$(model after attack) and Backdoor Accuracy($M'$) is the accuracy of $M'$ on poisoned test set points.}
\vspace{-1em}
\label{tab:nlp}
\end{table*}

%% file: cv_table.tex
\begin{table}[t]
\centering
\resizebox{\columnwidth}{!}{
\begin{tabular}{c|c|c|ccccc|}
\cline{2-8}
 & Test$(M)$     & $\epsilon$ & \multicolumn{1}{c}{Test$(M')$} & \multicolumn{1}{c}{Backdoor $(M')$} & \multicolumn{1}{c}{$\% \ \Delta \ell_{\infty}$} & \multicolumn{1}{c}{$\% \ \Delta \ell_{1}$} & $\% \ \Delta \ell_{2}$ \\ \hline
\multicolumn{1}{|c|}{\multirow{5}{*}{ResNet-20}} & \multirow{5}{*}{91.48} & 0.002                    & 86.82                                    & 18.11                                        & 0.09                                             & 2.54                                        & 1.76                   \\ \cline{3-8} 
\multicolumn{1}{|c|}{}                           &                        & 0.005                    & 87.27                                    & 90.62                                        & 0.23                                             & 5.96                                        & 4.20                   \\ \cline{3-8} 
\multicolumn{1}{|c|}{}                           &                        & 0.01                     & 89.76                                    & 99.78                                        & 0.46                                             & 9.19                                        & 6.88                   \\ \cline{3-8} 
\multicolumn{1}{|c|}{}                           &                        & 0.02                     & 90.03                                    & 99.95                                        & 0.91                                             & 10.46                                       & 8.69                   \\ \cline{3-8} 
\multicolumn{1}{|c|}{}                           &                        & $\infty$     &     90.21    &  99.98   &   2.19                   &    10.85   &  9.14     \\ \hline \hline
\multicolumn{1}{|c|}{\multirow{5}{*}{ResNet-32}} & \multirow{5}{*}{92.34} & 0.002                    & 88.25                                    & 36.78                                        & 0.11                                             & 3.24                                        & 2.24                   \\ \cline{3-8} 
\multicolumn{1}{|c|}{}                           &                        & 0.005                    & 90.43                                    & 99.42                                        & 0.27                                             & 6.59                                        & 4.80                   \\ \cline{3-8} 
\multicolumn{1}{|c|}{}                           &                        & 0.01                     & 91.48                                    & 99.96                                        & 0.55                                             & 9.73                                        & 7.56                   \\ \cline{3-8} 
\multicolumn{1}{|c|}{}                           &                        & 0.02                     & 91.52                                    & 99.95                                        & 1.10                                             & 11.34                                       & 9.45                   \\ \cline{3-8} 
\multicolumn{1}{|c|}{}                           &                        & $\infty$   &  91.82   &   99.99   &    2.75  &   12.25    &    10.33         \\ \hline \hline
\multicolumn{1}{|c|}{\multirow{5}{*}{ResNet-56}} & \multirow{5}{*}{93.27} & 0.002                    & 89.34                                    & 75.39                                        & 0.09                                             & 4.18                                        & 2.90                   \\ \cline{3-8} 
\multicolumn{1}{|c|}{}                           &                        & 0.005                    & 91.95                                    & 99.92                                        & 0.22                                             & 7.54                                        & 5.65                   \\ \cline{3-8} 
\multicolumn{1}{|c|}{}                           &                        & 0.01                     & 92.23                                    & 99.99                                        & 0.44                                             & 12.17                                       & 9.49                   \\ \cline{3-8} 
\multicolumn{1}{|c|}{}                           &                        & 0.02                     & 92.52                                    & 99.98                                        & 0.87                                             & 14.20                                       & 11.78                  \\ \cline{3-8} 
\multicolumn{1}{|c|}{}                           &                        & $\infty$   &   92.89   &   99.99   &  3.09 & 15.53  &  13.02  \\ \hline
\end{tabular}
}
\vspace{1em}
\caption{Adversarial weight perturbation for CIFAR-10 classification. Test($M$/$M'$) is the test set accuracy of $M$(original base model)/$M'$(model after attack) and Backdoor($M'$) is the accuracy of $M'$ on poisoned test set points.}
\label{tab:cv}
\end{table}
\setlength{\textfloatsep}{0pt}